\documentclass[letterpaper, 10 pt, conference]{ieeeconf}  
\IEEEoverridecommandlockouts                              

\usepackage{graphics} 
\usepackage{epsfig} 
\usepackage[ruled,vlined,linesnumbered,algosection,english,commentsnumbered]{algorithm2e}
\SetKwInput{KwInput}{Input}
\SetKwInput{KwOutput}{Output}
\usepackage{algpseudocode}
\usepackage{stfloats}
\usepackage{times} 
\usepackage{amsmath} 
\usepackage{amssymb}  
\usepackage[hang]{subfigure} 
\usepackage{cite}
\usepackage{siunitx}
\usepackage{soul}
\usepackage{color}
\usepackage{xcolor}
\usepackage{mathrsfs}
\usepackage[font=scriptsize]{caption}

\soulregister\ref7  
\soulregister\cite7 
\soulregister\eqref7
\soulregister\prettyref7

\usepackage{multirow}
\usepackage{algpseudocode}
\usepackage{makecell}
\usepackage{booktabs}
\usepackage{threeparttable}
\usepackage{hyperref}
\usepackage{endnotes}

\bstctlcite{IEEEexample:BSTcontrol}

\usepackage{prettyref}
\newrefformat{fig}{Fig.~\ref{#1}}
\newrefformat{sec}{Sec.~\ref{#1}}
\newrefformat{tab}{Tab.~\ref{#1}}
\newrefformat{algo}{Algo.~\ref{#1}}
\newrefformat{eq}{(\ref{#1})}

\usepackage{color}

\title{\LARGE \bf
  \textcolor{blue}{LiPS}: Large-Scale Humanoid Robot Reinforcement \textcolor{blue}{L}earning w\textcolor{blue}{i}th \textcolor{blue}{P}arallel-\textcolor{blue}{S}eries Structures}

\author{Qiang Zhang$^{1,2,\ast}$,  Gang Han$^{1,\ast,\clubsuit}$, Jingkai Sun$^{1,2,\ast}$, Wen Zhao$^{1,\ast}$,\\ Jiahang Cao$^{2}$, Jiaxu Wang$^{2}$, Hao Cheng$^{2}$, Lingfeng Zhang$^{2}$, Yijie Guo$^{1,\dagger}$, Renjing Xu$^{2, \dagger}$}

\begin{document}
\maketitle

\footnotetext[1]{The authors are with Beijing Innovation Center of Humanoid Robotics Co. Ltd. {\tt\small Jony.Zhang@xhumanoid.com}}
\footnotetext[2]{The authors are with The Hong Kong University of Science and Technology (Guangzhou), China. {$^{\ast}$ are equal contributions. $^{\dagger}$ are corresponding authors. $^{\clubsuit}$ is the project leader.} {\tt\small  \ qzhang749@connect.hkust-gz.edu.cn}}

\thispagestyle{empty}
\pagestyle{empty}

\begin{abstract}

In recent years, research on humanoid robots has garnered significant attention, particularly in reinforcement learning based control algorithms, which have achieved major breakthroughs. Compared to traditional model-based control algorithms, reinforcement learning based algorithms demonstrate substantial advantages in handling complex tasks. Leveraging the large-scale parallel computing capabilities of GPUs, contemporary humanoid robots can undergo extensive parallel training in simulated environments.
A physical simulation platform capable of large-scale parallel training is crucial for the development of humanoid robots.
As one of the most complex robot forms, humanoid robots typically possess intricate mechanical structures, encompassing numerous series and parallel mechanisms. However, many reinforcement learning based humanoid robot control algorithms currently employ open-loop topologies during training, deferring the conversion to series-parallel structures until the sim2real phase. This approach is primarily due to the limitations of physics engines, as current GPU-based physics engines often only support open-loop topologies or have limited capabilities in simulating multi-rigid-body closed-loop topologies.
For enabling reinforcement learning-based humanoid robot control algorithms to train in large-scale parallel environments, we propose a novel training method — \textcolor{blue}{LiPS}. By incorporating multi-rigid-body dynamics modeling in the simulation environment, we significantly reduce the sim2real gap and the difficulty of converting to parallel structures during model deployment, thereby robustly supporting large-scale reinforcement learning for humanoid robots.

\end{abstract}

\section{Introduction}

In recent years, humanoid robots have garnered increasing attention and are regarded as an ultimate robotic solution capable of perfectly adapting to human society and various work scenarios~\endnote{\href{https://agilityrobotics.com/robots}{https://agilityrobotics.com/robots}}~\endnote{\href{https://www.figure.ai/}{https://www.figure.ai/}}~\cite{noreils2024humanoid}. Researchers and society hold high expectations for them. Particularly, when combined with deep learning-based artificial intelligence, their potential becomes even greater~\cite{zhang2024whole}~\cite{wei2023learning}~\cite{zhang2024wococo}~\cite{fu2024humanplus}~\cite{cheng2024expressive}~\cite{zhuang2024humanoid}. Deep learning has achieved significant breakthroughs in visual and language tasks, and humanoid robots, due to their anthropomorphic form, possess unique advantages among all types of robots. This anthropomorphic characteristic enables humanoid robots to adapt to human living environments and directly utilize human data for imitation learning and training.

\begin{figure}[h]
    \centering
    \includegraphics[width=\linewidth]{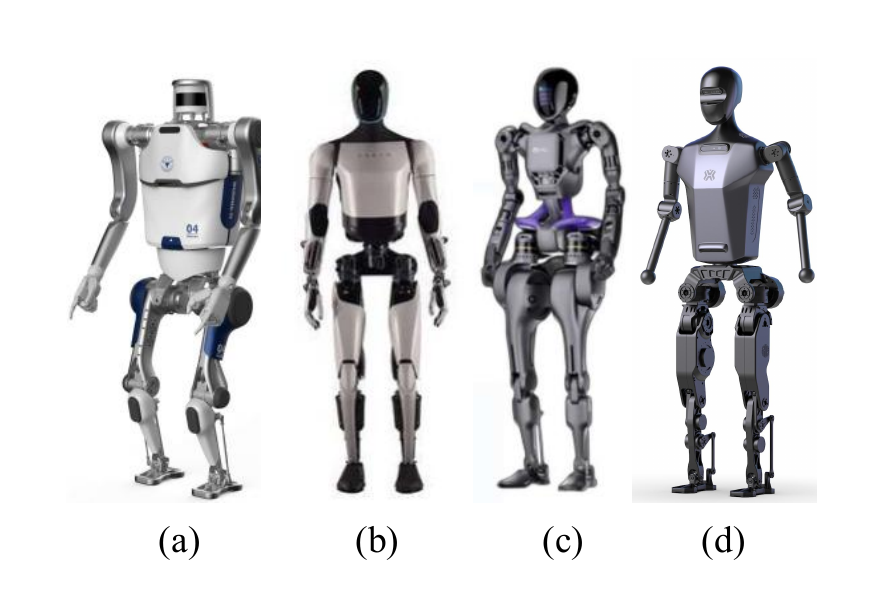}
    \caption{Four well-known humanoid robots all use parallel ankle mechanisms, but each has a different design approach: (a) Wukong uses a nearly decoupled parallel structure, (b) Tesla Optimus employs linear actuators for the parallel ankle design, while (c) Fourier GR-1 and (d) Tien Kung use rotary actuators but have different structural approaches at the ankle connection.}
    \label{fig:Presentation1}
    \vspace{-2mm}
\end{figure}

Humanoid robots are an extremely complex form of robotics. Typically, humanoid robots have intricate bipedal and bimanual mechanical structures, which include numerous series and parallel mechanisms. This complexity makes the control problem of humanoid robots a typical nonlinear, multivariable, and time-varying control problem. Traditional model-based control algorithms often require precise modeling of the robot and its operating environment, resulting in highly complex models. Additionally, these models usually need to be readjusted when generalized to different tasks and environments. Therefore, an increasing number of researchers are adopting deep reinforcement learning methods to address the control problems of humanoid robots.
We present four renowned humanoid robots in Fig.\ref{fig:Presentation1}, including Tien Kung, Optimus, Wukong, and GR-1.


The training frameworks and simulation engines for deep reinforcement learning in robotics often require a trade-off between speed and accuracy. Common model file formats and physics engines each have their own advantages and disadvantages. URDF~\endnote{\href{https://wiki.ros.org/urdf}{https://wiki.ros.org/urdf}} (Unified Robot Description Format) is simple and easy to use, widely applied in the ROS~\endnote{\href{https://wiki.ros.org}{https://wiki.ros.org}}  ecosystem, and is currently the most widely used model file format. However, URDF's description of parallel structures neglects dynamic characteristics, retaining only the geometric characteristics of parallel structures, resulting in lower accuracy. SDF (Simulation Description Format) can detail the dynamic characteristics of parallel structures, but the model files are more complex and difficult to support large-scale parallel training. MuJoCo~\cite{todorov2012mujoco} XML can accurately describe the dynamic characteristics of parallel structures, but its physics engine does not support running on GPUs, making large-scale parallel training difficult. Finite element analysis-based physics engines provide very accurate simulation results but usually perform simulations through the CPU, resulting in very slow speeds and making large-scale parallel training infeasible. In contrast, GPU-based physics engines can perform large-scale parallel training, significantly improving training speed, but their accuracy is not as high as finite element analysis-based physics engines. By understanding the advantages and disadvantages of these model file formats and physics engines, we can better choose the simulation tools that suit specific application needs.

Currently, humanoid robot control algorithms typically employ open-loop topologies during training and complete the series-parallel conversion during the sim2real phase. Due to the complex structure and dynamic characteristics of humanoid robots, the training process is often much slower than that of quadruped robots, requiring neural networks more time to search for the optimal solution in the action space. This also means that reinforcement learning for humanoid robots demands higher parallel sampling efficiency. A significant amount of research on humanoid robots utilizes the GPU-based training environment IsaacGym~\cite{makoviychuk2021isaac} and describes robot configurations using the URDF file format. URDF is simple, easy to use, and widely adopted, making it a highly suitable model file format for most current work. Developing large-scale parallel training for more complex robot structures based on URDF is crucial for improving training efficiency and effectiveness.

We propose LiPS, a novel training method that optimizes the modeling of complex multi-rigid-body dynamics in the simulation environment, significantly reducing the difficulty of converting parallel structures and the sim2real gap during the sim2real phase. This method allows us to more effectively simulate the complex dynamic behaviors of humanoid robots, achieving more efficient and precise training in the simulation environment, ultimately improving performance and reliability in practical applications. Additionally, our method can be easily transferred to other URDF-based robots, demonstrating good generalization capabilities. It can assist the entire robotics research community in conducting large-scale reinforcement learning training for URDF-based robots.Our contributions can be summarized as follows:
\begin{enumerate}
    \item We propose LiPS, a novel training method that optimizes the modeling of complex multi-rigid-body dynamics in the simulation environment. This significantly reduces the difficulty of converting parallel structures and the sim2real gap during the sim2real phase for robots described by URDF model files. We provide the entire robotics research community with an effective tool for large-scale reinforcement learning training of complex robots based on URDF.
    \item LiPS enhances both training efficiency and the inference efficiency of robots in real-world applications. Our method simulates the complex dynamic behaviors of humanoid robots, reducing the computational load and errors during real-world deployment.
\end{enumerate}

\section{Related Work}
\subsection{Robot Reinforcement Learning Platforms}

In the field of robotics, simulation plays a crucial role in both AI learning-based and model-based approaches. Simulators allow researchers to test and validate algorithms in a virtual environment before deploying them on physical hardware, ensuring safety and accelerating development. This is particularly important for reinforcement learning (RL), which requires extensive trial and error to learn effective policies. Simulators can provide the necessary data for training without the high costs and risks associated with real-world experiments.

One of the most advanced platforms for robot reinforcement learning is NVIDIA's Isaac Gym~\cite{makoviychuk2021isaac}. Isaac Gym leverages GPU acceleration to perform high-fidelity physical simulations and neural network training simultaneously on the GPU, bypassing the CPU bottleneck and enabling rapid training. This platform supports parallel simulation of numerous environments, significantly enhancing training efficiency. Isaac Gym integrates the PhysX physics engine and RTX rendering engine, providing accurate physical simulations and realistic visualizations. 
Another notable platform is Webots~\endnote{\href{https://github.com/cyberbotics/webots}{https://github.com/cyberbotics/webots}} , an open-source robot simulator developed by Cyberbotics Ltd. Webots offers a comprehensive environment for robot development and modeling, with a physics engine based on ODE. It is widely used in both industrial and academic research due to its ease of use and extensive features.
CoppeliaSim~\endnote{\href{https://www.coppeliarobotics.com/}{https://www.coppeliarobotics.com/}} , formerly known as V-REP, is another popular simulator used in industrial, educational, and research settings. 
Gazebo~\endnote{\href{https://gazebosim.org/home}{https://gazebosim.org/home}}, an open-source simulator integrated with the ROS ecosystem, is extensively used in robotics research. 
PyBullet~\cite{coumans2021}, based on the Bullet physics engine, is another open-source simulator widely used in RL.

In addition to these established platforms, NVIDIA recently introduced IsaacLab~\cite{mittal2023orbit}, a new simulation environment designed to further enhance robot learning capabilities. IsaacLab builds upon the strengths of Isaac Gym, aiming to provide even more advanced features for robotics research and development. However, it's important to note that IsaacLab is still a relatively new environment, and its user base is currently limited. As with many emerging technologies, it is still under active development and may have some issues that are being addressed. Despite these challenges, IsaacLab shows promise in advancing the field of robot simulation and learning, and its development is being closely watched by the robotics community.


\subsection{Reinforcement Learning on Humanoid Robot}
Legged robot locomotion has made significant strides, serving as a foundational aspect of advancing humanoid robotics. The integration of Reinforcement Learning (RL) has been instrumental in these developments~\cite{lirobust, siekmann2021blind}. For instance, the robot Cassie has demonstrated a variety of walking and running patterns using periodic-parameterized reward functions~\cite{siekmann2021sim} and even set a Guinness World Record for the fastest 100-meter dash by employing pre-optimized gaits~\cite{crowley2023optimizing}.~\cite{liao2024berkeley} develop a humanoid research platform for learning-based locomotion tasks.~\cite{shi2022reference} expanded the capabilities of legged robots by introducing an assistive force curriculum, allowing for agile motion learning even in environments without explicit reference trajectories.  ~\cite{kim2023torque} proposed a torque-based approach that effectively bridges the gap between simulated training and real-world application, further advancing the field. DeepMind's innovative research~\cite{haarnoja2023learning} enabled a miniature humanoid robot to acquire complex soccer skills through a combination of teacher-student distillation and self-play techniques. Moreover, the use of attention-based transformers in the Digit humanoid robot has resulted in more adaptable and versatile locomotion patterns~\cite{radosavovic2023learning}. Most recently, ~\cite{zhang2024whole} employed an adversarial training method, which makes more human-like and higher-performing locomotion in real-world robot experiments.

\section{Preliminary}

\subsection{Humanoid Robots Based on Complex Series-Parallel Structures}

\begin{figure}[h]
    \centering
    \includegraphics[width=\linewidth]{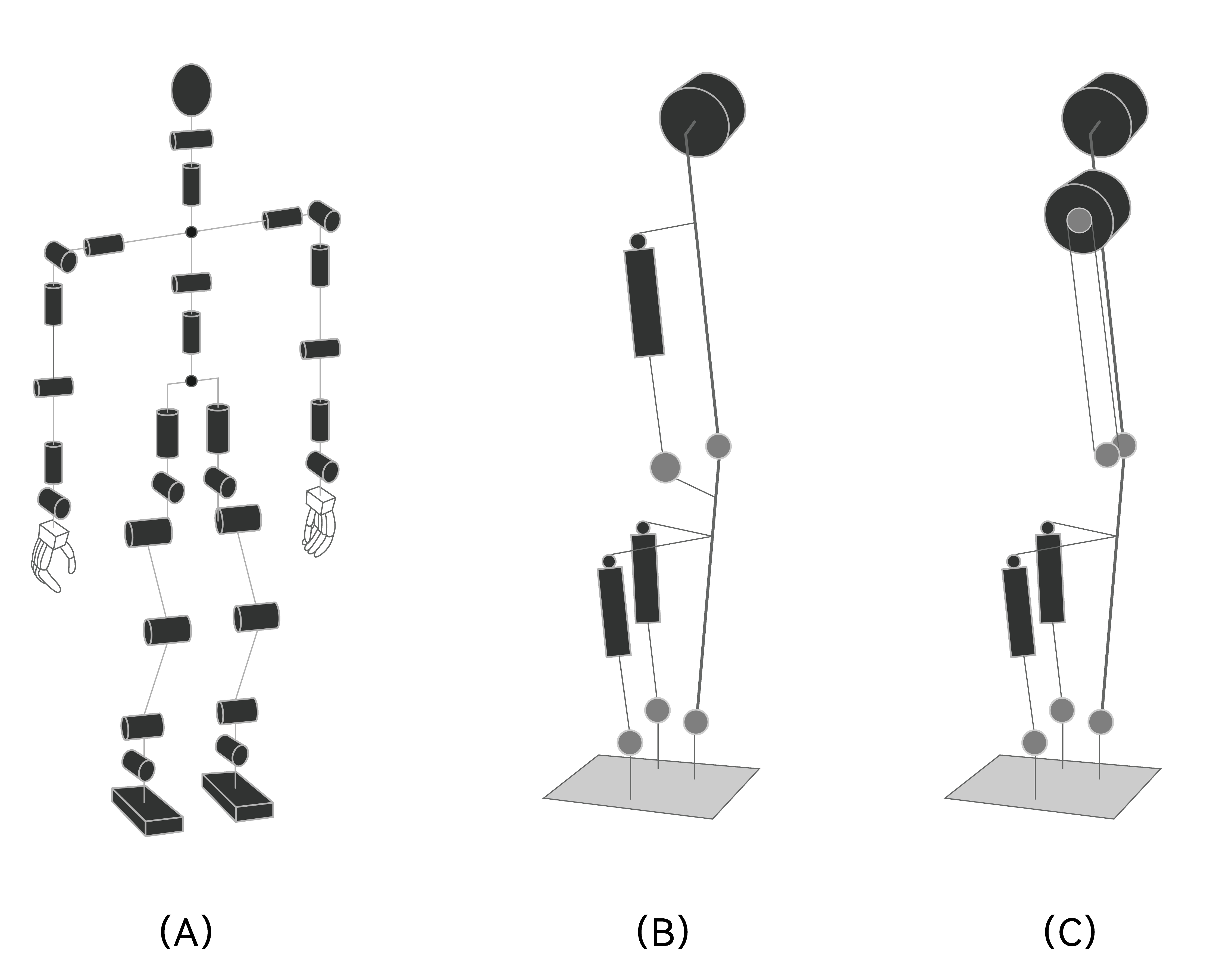}
    \caption{We referred to the description of current humanoid robot configurations in ~\cite{noreils2024humanoid} and ~\cite{muhammad2010closed}, similarly conducted a detailed analysis of the ankle structure.}
    \label{fig:structure}
\end{figure}

The design and control of humanoid robots have long been significant challenges in the field of robotics. In recent years, humanoid robot designs based on complex series-parallel structures have been leading innovation in this domain. Traditional humanoid robots typically employed simple serial structures, with rotary joints directly driven by servomotors in a serial configuration (Fig.~\ref{fig:structure}). While this structure facilitated modeling and control, it exhibited notable limitations in dynamic performance and efficiency. As research progressed, engineers gradually recognized that complex mechanical designs integrating both serial and parallel structures could significantly enhance robot performance.

Particularly noteworthy is the innovation in ankle structure. Almost all new-generation humanoid robots have adopted complex parallel structures in their ankles, typically using two actuators through a universal joint to achieve pitch and roll control (Fig.\ref{fig:structure} (B)(C)). This design not only enhances the ankle's stiffness and load-bearing capacity but also significantly improves the robot's ability to adapt to uneven terrain and maintain dynamic balance. However, these complex parallel structures also present new challenges, especially in dynamic modeling and control algorithm design. Traditional model-based control methods often struggle to fully leverage the advantages of these structures, while learning-based approaches, such as reinforcement learning, have shown tremendous potential.


\section{Method}
\subsection{Dynamics Modeling of Humanoid Robots with Multi-Rigid-Body and Floating Base}

\begin{figure}[h]
    \centering
    \includegraphics[width=0.6\linewidth]{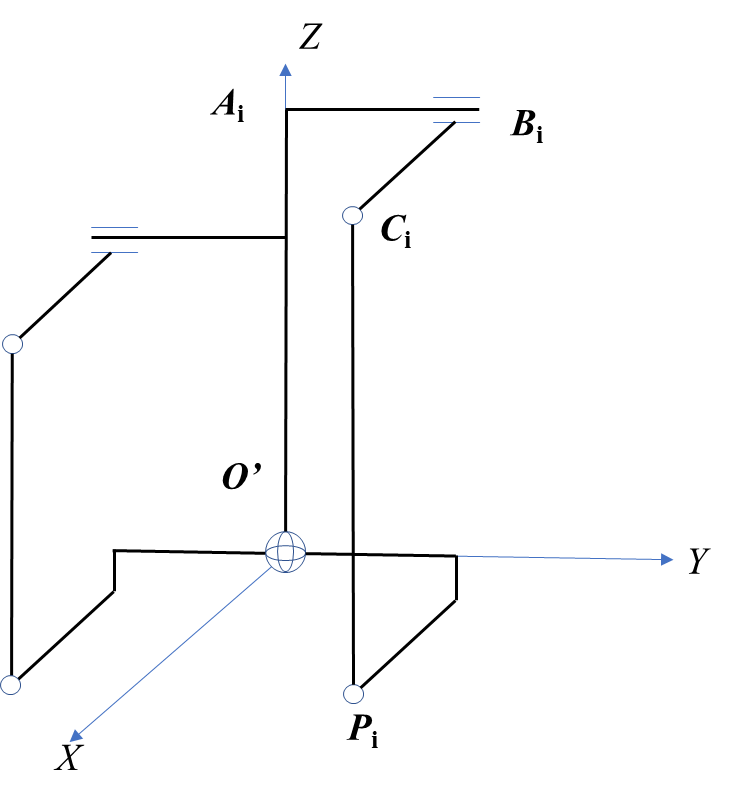}
    \caption{Schematic Diagram of Ankle Dynamics Modeling}
    \label{fig:ik}
\end{figure}

Humanoid robots primarily exhibit the following characteristics: multi-rigid-body structure, floating base, and parallel ankle mechanism. Based on these features, we model the local kinematics of the closed-loop linkage parallel ankle and the dynamics of the multi-rigid-body floating base. We begin by constructing the local kinematics of the closed-loop linkage parallel ankle.

Using the coordinate system fixed to the lower leg as the reference, we can analyze the kinematics as in Fig.~\ref{fig:ik}:
If the posture of the end ankle relative to the lower leg is given as $\chi=[\phi,\theta]^T$, the position of the hinge point $P_{i}$ of the parallel linkage relative to the center of the footplate $\mathbf{o}^{\prime}$ in the lower leg coordinate system can be expressed as:

\begin{equation}
    \boldsymbol{o}’\boldsymbol{P}_i=\boldsymbol{R}_Y(\theta)\boldsymbol{R}_X(\phi)\boldsymbol{o}’\boldsymbol{P}_i’
\end{equation}

\begin{equation}
    \boldsymbol{R}_X(\phi)=\begin{bmatrix}
    1 & 0 & 0 \\
    0 & \cos\phi & -\sin\phi \\
    0 & \sin\phi & \cos\phi
    \end{bmatrix}
\end{equation}
\begin{equation}
    \boldsymbol{R}_Y(\theta)=\begin{bmatrix}
    \cos\theta & 0 & \sin\theta \\
    0 & 1 & 0 \\
    -\sin\theta & 0 & \cos\theta
    \end{bmatrix}
\end{equation}

The distance between the end point of the active arm of the parallel joint and the hinge point of the parallel linkage always satisfies the length of the passive rod $L_2$. $L_1$ is the length of the active arm. Assuming the current position of the active arm is at the zero position, the position of the point relative to the center of the footplate is:
\begin{equation}
    \boldsymbol{o}^{\prime}C_1=[L_1\cos q_1,r_1,o^{\prime}A_1-L_1\sin q_1]
\end{equation}
\begin{equation}
    \boldsymbol{o}^{\prime}C_2=[L_1\cos q_2,-r_1,o^{\prime}A_2-L_1\sin q_2]
\end{equation}
The main joint angles of the robot can be obtained using the vector loop method:

\begin{align}
    2\left(\boldsymbol{o}'\boldsymbol{P}_1\right)_x L_1 \cos q_1 
    &- 2\left(\left(\boldsymbol{o}'\boldsymbol{P}_1\right)_z - \boldsymbol{o}'A_1\right) L_1 \sin q_1 \notag \\
    &= \left(\boldsymbol{o}'\boldsymbol{P}_1\right)_x^2 
    + \left(\left(\boldsymbol{o}'\boldsymbol{P}_1\right)_y - \boldsymbol{r}_1\right)^2 \notag \\
    &\quad + \left(\left(\boldsymbol{o}'\boldsymbol{P}_1\right)_z - \boldsymbol{o}'A_1\right)^2 
    + L_1^2 - L_2^2 \\
    2\left(\boldsymbol{o}'\boldsymbol{P}_2\right)_x L_1 \cos q_2 
    &- 2\left(\left(\boldsymbol{o}'\boldsymbol{P}_2\right)_z - \boldsymbol{o}'A_2\right) L_1 \sin q_2 \notag \\
    &= \left(\boldsymbol{o}'\boldsymbol{P}_2\right)_x^2 
    + \left(\left(\boldsymbol{o}'\boldsymbol{P}_2\right)_y + \boldsymbol{r}_1\right)^2 \notag \\
    &\quad + \left(\left(\boldsymbol{o}'\boldsymbol{P}_2\right)_z - \boldsymbol{o}'A_2\right)^2 
    + L_1^2 - L_2^2
\end{align}

Needs to satisfy the constraints:
\begin{equation}
    a_i^2+b_i^2-c_i^2\geq0,i=1,2 , -1\leq\cos q_i\leq1
\end{equation}

Four solutions are obtained, and the final result is selected based on the constraints of the rotation range and the length of the rod $L2$. 
Based on the position inverse solution, further analysis of the inverse solution for the velocity and acceleration of the parallel ankle mechanism is conducted.
According to the position inverse solution, the passive linkage vector $C_iP_i$, the active drive arm vector $B_iC_i$, and the relative position of the drive shaft point $B_i$ can be determined.

Let $\dot{\chi}=[\dot{\phi},\dot{\theta}]^T$ and $\ddot{\chi}=[\ddot{\phi},\ddot{\theta}]^T$ represent the velocity and acceleration of the footplate, respectively.
The velocity of point $P_i$ a in the base coordinate system $\{A\}:o’-XYZ$ is expressed as follows:

\begin{equation}
    ^A\boldsymbol{V}_{P_i}={}^A\boldsymbol{R}_{P_i}\dot{\boldsymbol{\chi}}=\begin{bmatrix}-\boldsymbol{o}^{\prime}\boldsymbol{P}_i\end{bmatrix}_\times\boldsymbol{R}_\omega\dot{\boldsymbol{\chi}}
\end{equation}
$^A\boldsymbol{R}_{P_i}$ can be expressed as:
\begin{equation}
    \begin{aligned}
        \begin{bmatrix}
            0 & \boldsymbol{R}_{XY}(2,2)\big(\boldsymbol{o}'P_i\big)_z - \boldsymbol{R}_{XY}(3,2)\big(\boldsymbol{o}'P_i\big)_y \\
            -\big(o'P_i\big)_z & -\boldsymbol{R}_{XY}(1,2)\big(\boldsymbol{o}'P_i\big)_z + \boldsymbol{R}_{XY}(3,2)\big(\boldsymbol{o}'P_i\big)_x \\
            \big(o'P_i\big)_y & \boldsymbol{R}_{XY}(1,2)\big(\boldsymbol{o}'P_i\big)_y - \boldsymbol{R}_{XY}(2,2)\big(\boldsymbol{o}'P_i\big)_x
        \end{bmatrix}
    \end{aligned}
\end{equation}

$W_{1i}$ is the angular velocity of the active driving arm: $W_{1i}=(0,\dot{q}_i,0)^T$.
The velocity of point $C_i$ can be expressed by the rotational speed of the active arm.
\begin{equation}
    \boldsymbol{W}_{1i}\times\boldsymbol{B}_i\boldsymbol{C}_i={}^A\boldsymbol{R}_{C_i}\dot{\boldsymbol{\chi}},{}^A\boldsymbol{R}_{C_i}=\begin{bmatrix}\left(\boldsymbol{B}_i\boldsymbol{C}_i\right)_z\\0\\-\left(\boldsymbol{B}_i\boldsymbol{C}_i\right)_x\end{bmatrix}\boldsymbol{J}(i,:)
\end{equation}

The variable $\dot{q}_i$ represents the rotational speeds of the two main joints. Based on the principle that the projection of the velocities of the two endpoints of the fixed-length passive rod along the rod is equal, we can derive the formula for $\dot{q}_i$. The rotational speeds of the two main joints can be expressed as:

\begin{equation}
\dot{\boldsymbol{q}} = [\dot{q}_1, \dot{q}_2]^T = \boldsymbol{J} \dot{\boldsymbol{\chi}},\! \boldsymbol{J}(i, :) = \frac{1}{\left(\boldsymbol{B}_i \boldsymbol{C}_i \times \boldsymbol{C}_i \boldsymbol{P}_i\right)_y} \boldsymbol{C}_i \boldsymbol{P}_i^T {}^A \boldsymbol{R}_{P_i}
\end{equation}
Thus, $\boldsymbol{J}$ is the inverse Jacobian matrix of this high-speed parallel robot, thereby completing the derivation of the inverse velocity solution.
The acceleration of point $P_i$ in the base coordinate system $\{A\}:o-XYZ$:

\begin{align}
    {}^{A}\dot{\boldsymbol{V}}_{{P_{i}}} &= {}^{A}\boldsymbol{R}_{{P_{i}}}\ddot{\boldsymbol{\chi}} + {}^{A}\dot{\boldsymbol{R}}_{{P_{i}}}\dot{\boldsymbol{\chi}}, \\
    {}^{A}\dot{\boldsymbol{R}}_{{P_{i}}}\dot{\boldsymbol{\chi}} &= -\boldsymbol{o}^{\prime}\boldsymbol{P}_{i}\times\left(\dot{\boldsymbol{R}}_{\omega}\dot{\boldsymbol{\chi}}\right) - \left({}^{A}\boldsymbol{R}_{{P_{i}}}\dot{\boldsymbol{\chi}}\right)\times\left(\boldsymbol{R}_{\omega}\dot{\boldsymbol{\chi}}\right), \\
    {}^{A}\dot{\boldsymbol{R}}_{{P_{i}}} &= -\left[\boldsymbol{o}^{\prime}\boldsymbol{P}\right]_{{\mathbf{x}}}\dot{\boldsymbol{R}}_{\omega} - \left[{}^{A}\boldsymbol{R}_{{P_{i}}}\dot{\boldsymbol{\chi}}\right]_{{\mathbf{x}}}\boldsymbol{R}_{\omega}
\end{align}

where $\boldsymbol{R}$ is the projection matrix of the rotational speed of $\dot{\chi}=[\dot{\phi},\dot{\theta}]^T$ and the footplate relative to the lower leg coordinate system.
The acceleration of $P_i$ and $C_i$ projected along the passive rod satisfies the following equation:
\begin{multline}
    \left(\boldsymbol{B}_{i}\boldsymbol{C}_{i}\boldsymbol{\times}\boldsymbol{C}_{i}\boldsymbol{P}_{i}\right)^{T}\dot{\boldsymbol{W}}_{1i} 
    - \boldsymbol{W}_{1i}{}^2\boldsymbol{C}_{i}\boldsymbol{P}_{i}^{T}\boldsymbol{B}_{i}\boldsymbol{C}_{i} = \\
    \boldsymbol{C}_{i}\boldsymbol{P}_{i}^{T}\left({}^{A}\boldsymbol{R}_{{P_{i}}}\ddot{\boldsymbol{\chi}}+{}^{A}\dot{\boldsymbol{R}}_{{P_{i}}}\dot{\boldsymbol{\chi}}\right) + \\
    \dot{\boldsymbol{\chi}}^{T}\left({}^{A}\boldsymbol{R}_{{P_{i}}}-{}^{A}\boldsymbol{R}_{{C_{i}}}\right)^{T}\left({}^{A}\boldsymbol{R}_{{P_{i}}}-{}^{A}\boldsymbol{R}_{{C_{i}}}\right)\dot{\boldsymbol{\chi}}
\end{multline}

Thus, the angular acceleration of the $i$ primary joint can be expressed in the following form:
\begin{multline}
    \left(\boldsymbol{B}_{i}\boldsymbol{C}_{i}\boldsymbol{\times}\boldsymbol{C}_{i}\boldsymbol{P}_{i}\right)^{T}\dot{\boldsymbol{W}}_{1i}= \boldsymbol{C}_{i}\boldsymbol{P}_{i}^{T}{}^{A}\boldsymbol{R}_{{P_{i}}}\ddot{\boldsymbol{\chi}} \\ +\biggl[\boldsymbol{C}_{i}\boldsymbol{P}_{i}^{T}({}^{A}\dot{\boldsymbol{R}}_{{P_{i}}}+\boldsymbol{B}_{i}\boldsymbol{C}_{i}\dot{\boldsymbol{\chi}}^{T}\boldsymbol{J}^{T}{}_{(i,:)}\boldsymbol{J}_{(i,:)}) \\+ \dot{\boldsymbol{\chi}}^{T}({}^{A}\boldsymbol{R}_{{P_{i}}}-{}^{A}\boldsymbol{R}_{{C_{i}}})^{T}({}^{A}\boldsymbol{R}_{{P_{i}}}-{}^{A}\boldsymbol{R}_{{C_{i}}})\biggr]\dot{\boldsymbol{\chi}}
\end{multline}

Where $\boldsymbol{J}_{(i,:)}$ represents the \( i \)-th row vector of $\boldsymbol{J}$. Using the above method, the angular accelerations of the four main joints can be expressed in the following form, and the expression for $\boldsymbol{\dot{J}}$ can be obtained.

\begin{equation}
    \ddot{\boldsymbol{q}}=[\ddot{q}_1,\ddot{q}_2,\ddot{q}_3,\ddot{q}_4]^T=\boldsymbol{J}\ddot{\boldsymbol{\chi}}+\dot{\boldsymbol{J}}\dot{\boldsymbol{\chi}}
\end{equation}
\vspace{-6mm}

\begin{multline}
    \vspace{-6mm}
    \dot{\boldsymbol{J}}(i,:)=\frac1{\left(\boldsymbol{B}_i\boldsymbol{C}_i\boldsymbol{\times}\boldsymbol{C}_i\boldsymbol{P}_i\right)_y}\biggl[\boldsymbol{C}_i\boldsymbol{P}_i^T(^A\dot{\boldsymbol{R}}_{P_i}+ \\ \boldsymbol{B}_i\boldsymbol{C}_i\dot{\boldsymbol{\chi}}^T\boldsymbol{J}_{(i,:)}^T\boldsymbol{J}_{(i,:)})+\dot{\boldsymbol{\chi}}^T(^A\boldsymbol{R}_{P_i}-^A\boldsymbol{R}_{C_i})^T(^A\boldsymbol{R}_{P_i}-^A\boldsymbol{R}_{C_i})\biggr]
\end{multline}

At this point, we can obtain the acceleration of point $C_i$ in the base coordinate system $\{A\}:o-XYZ$. 

\begin{equation}
    \begin{aligned}&^A\dot{\boldsymbol{V}}_{C_i}={}^A\boldsymbol{R}_{C_i}\ddot{\boldsymbol{\chi}}+{}^A\dot{\boldsymbol{R}}_{C_i}\dot{\boldsymbol{\chi}},\\&^A\dot{\boldsymbol{R}}_{C_i}=\begin{bmatrix}\left(\boldsymbol{B}_i\boldsymbol{C}_i\right)_z\\-\left(\boldsymbol{B}_i\boldsymbol{C}_i\right)_x\end{bmatrix}\dot{\boldsymbol{J}}(i,:)-\boldsymbol{B}_i\boldsymbol{C}_i\dot{\boldsymbol{\chi}}^T\boldsymbol{J}(i,:)^T\boldsymbol{J}(i,:)\end{aligned}
\end{equation}
The inverse kinematic model for the acceleration of each major node has been established. The above acceleration is relative to the lower leg coordinate system.

\vspace{-6mm}
\subsection{Formulation in Large-Scale Parallel Reinforcement Learning Environment}
\vspace{-6mm}

\begin{figure}[h]
    \centering
    \includegraphics[width=\linewidth]{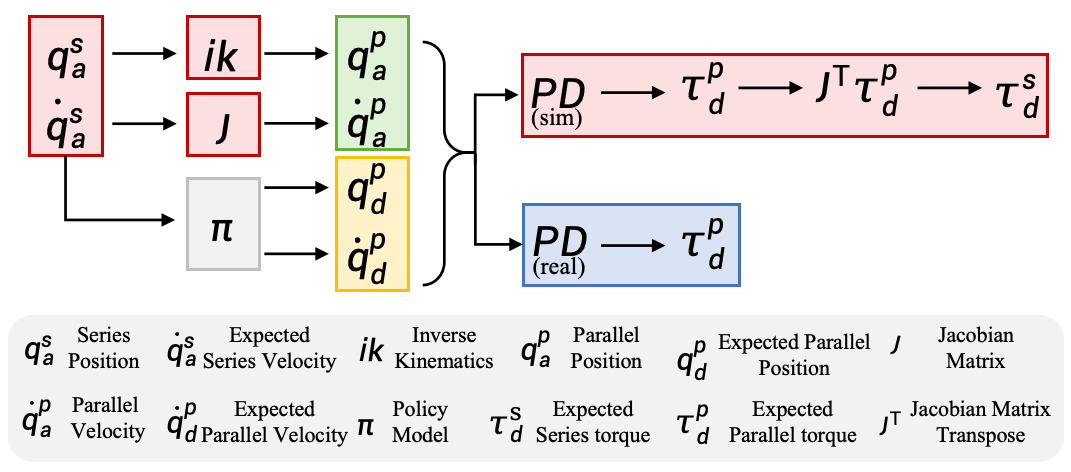}
    \caption{Illustration of LiPS Simulation Training and Real-World Deployment Process.}
    \label{fig:p2s}
\end{figure}

We demonstrate two methods for handling humanoid robot control problems with parallel structures (such as parallel ankles): the LiPS method and the currently widely used method. These two methods differ in their approach to training and deployment stages.

\textit{LiPS Method (Our Proposed Method):}
The LiPS method directly uses the parallel model for training and deployment in the reinforcement learning environment. As shown in Algorithm~\ref{alg:lips}, this method considers the dynamic characteristics of the parallel structure during the training process, including computing parallel generalized coordinates, dynamics matrices, and external force Jacobian matrices. The trained policy can be directly applied to the physical robot without additional conversion steps. The specific dynamics model and computation process of LiPS have been described in detail in the previous section.

\begin{algorithm}
\caption{LiPS: RL Training and Deployment with Parallel Ankle Mechanism}
\label{alg:lips}
\SetAlgoLined
\KwInput{series robot model, Environment parameters, RL algorithm}
\KwOutput{Trained policy for humanoid robot control}
Initialize robot model with series ankle mechanism\;
Initialize RL environment with series model\;
\While{not converged}{
Observe current state $s_t$ from environment\;
$a_t(Parallel) \leftarrow \text{RLPolicy}(s_t(Series))$\;
$\boldsymbol{\tau}_{\mathrm{d}}^{\boldsymbol{p}} \leftarrow \text{PD Controller}(a_t(Parallel))$\;
$J_{transpose} \leftarrow \text{ComputeTransposedJacobian}$\;
$\boldsymbol{\tau}_{\mathrm{d}}^{\boldsymbol{s}} \leftarrow \text{ComputeSeriesDesiredTorque}(J_{transpose}\boldsymbol{\tau}_{\mathrm{d}}^{\boldsymbol{p}})$\;
Apply $\tau_{Series}$ to the environment\;
Observe next state $s_{t+1}$ and reward $r_t$\;
Update RL policy using $(s_t, a_t, r_t, s_{t+1})$\;
}
// Deployment on physical robot
$\text{TrainedPolicy} \leftarrow \text{Trained RL policy}$\;
\While{robot is operating}{
Observe current state $s_t$\;
$a_t \leftarrow \text{TrainedPolicy}(s_t)$\;
Apply $a_t$ ($\tau_{parallel}$) to physical robot\;
}
\Return{Deployed policy on physical robot}
\hrule 
\textbf{Variable Definitions:}
\begin{itemize}
    \item $s_t, s_{t+1}$: Current and next state
    \item $a_t$: Action from RL policy
    \item $r_t$: Reward
    \item The other parameters are consistent with those shown in Fig.\ref{fig:p2s}
\end{itemize}

\end{algorithm}

\textit{Currently Widely Used Method:}
The currently widely used method treats the robot as a serial structure during the reinforcement learning training stage and only performs serial-to-parallel conversion when deploying to the physical robot. This method uses a simplified serial model during the training process, but requires additional conversion steps during the deployment stage to transform the output of the serial model into control signals suitable for the parallel structure.


The main advantage of the LiPS method (Fig.\ref{fig:p2s}) is that it considers the complex dynamics of the parallel structure during the training phase, thereby reducing errors in sim-to-real conversion. This method can more accurately simulate the actual behavior of the robot, contributing to learning more robust and efficient control strategies.
In contrast, the currently widely used method uses a simplified serial model during the training phase, which has a lower computational burden, but requires additional conversion during the deployment phase. This method may lead to larger differences between the training environment and the actual environment, increasing the difficulty of sim-to-real conversion.

By comparing these two methods, we can see that the LiPS method has potential advantages in handling humanoid robots with complex parallel structures, especially in reducing the sim-to-real gap and improving control accuracy.

\section{Experiments}
\subsection{Training and Implementation Details}
Our training was conducted on a single NVIDIA 4090 GPU, utilizing 4096 parallel environments in IsaacGym for each training session. This setup allowed for efficient and scalable learning of our reinforcement learning policy. The use of multiple environments enabled the collection of diverse experiences, contributing to the robustness and generalization capabilities of our trained model. 

We implemented and validated our approach through real-world experiments on the Tien Kung humanoid robot, developed by the Beijing Innovation Center of Humanoid Robotics Co. Ltd. Tien Kung is a sophisticated humanoid platform standing 163 cm tall and weighing 56 kg, equipped with 42 degrees of freedom. It features multiple visual perception sensors, six-dimensional force sensors, an inertial measurement unit (IMU), and 3D vision sensors. With a computational power of 550 trillion operations per second, Tien Kung provided an ideal testbed for our LiPS framework. For real-time deployment, we utilized an NVIDIA Orin for model inference, maintaining a control frequency of 100 Hz, which ensured smooth and responsive operation of the robot during task execution.

We used Legged Gym~\cite{rudin2022learning} as our codebase and integrated our series-parallel training module into the simulation. We adopted the same network training parameters as Legged Gym. No historical data or pre-trained models were used. We employed an MLP network with sizes [512, 256, 256] to test our results.

\subsection{Comparing Other Series-Parallel Approaches}

\begin{figure}[h]
    \centering
    \includegraphics[width=\linewidth]{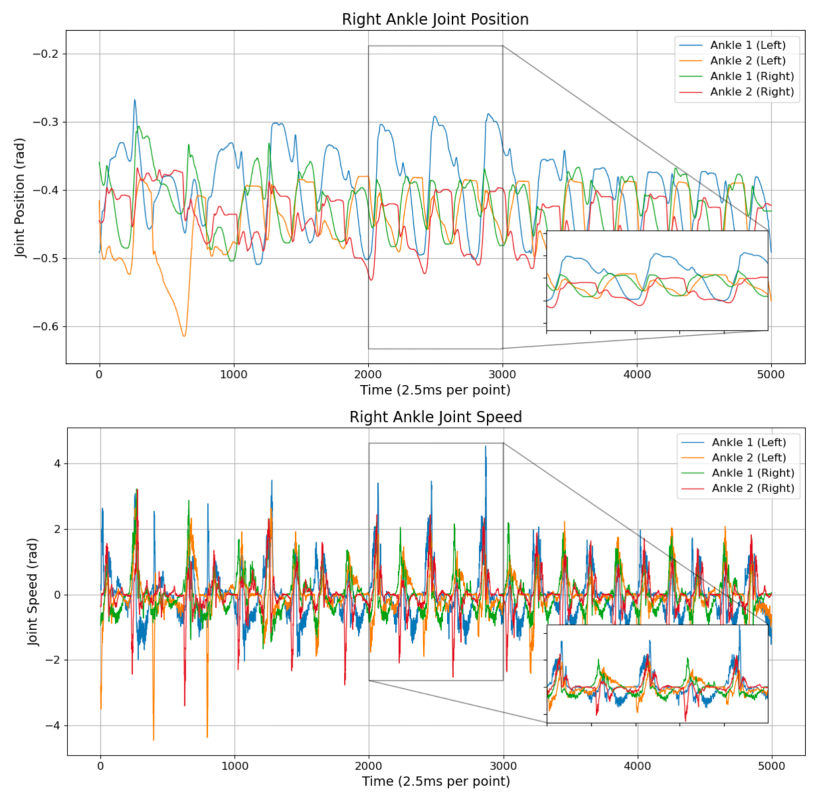}
    \caption{Joint velocities and positions of the two joints in the parallel structure of the ankle during the actual operation of the robot.}
    \label{fig:wave}
\end{figure}

In this paper, we compare three methods: (1) Using series-parallel conversion during deployment on actual robots, (2) Using passive ankles, and (3) Our LiPS method.

(1) Gu et al.~\cite{gu2024advancing} proposed two methods using active and passive ankles. The active ankle involves setting the ankle joint with corresponding kp and kd to achieve a closed kinematic chain ankle mechanism. However, the paper does not detail the specific methods and techniques used. Series-parallel conversion has been extensively studied in mechanical structure research~\cite{yoshikawa1985dynamic,van1999efficient,liu2018novel,han2018evaluation,han2020technology}. Here, we discuss two common approaches:
   ~\textbf{a. Using torque equivalence.} As shown in Fig.~\ref{fig:p2s}, the implementation in simulation is similar. Since we use serial mechanisms in training and model description, we can only obtain the desired positions and velocities of the joints in the serial mechanism. We can convert the joint torques from the serial mechanism to the parallel mechanism using analytical solutions. The problem with this method is that solving the analytical solutions at the same frequency as the joint control on the robot side consumes a lot of computational resources and may result in no solutions.
   ~\textbf{b. Using position equivalence.} Directly use the joint positions in the parallel mechanism as the desired positions and control the joint positions in the parallel mechanism using a PD controller. This can be simply approximated by treating the two parallel joints as equivalent average forces. The problem with this method is that if the encoder precision at the joint side is poor, the error will be amplified, causing the robot to jitter. We show the jitter and spikes of a high-performance joint actuator during robot movement in Fig.~\ref{fig:p2s}. Additionally, the simple and crude equivalent average force cannot fully utilize the design performance of the parallel ankle.

\begin{table}[h]
\centering
\begin{tabular}{|c|c|c|c|}
\hline
\textbf{Different} & \textbf{Computational} & \textbf{Control} & \textbf{Noise} \\
\textbf{Method} & \textbf{Load} & \textbf{Performance} & \textbf{Robustness} \\
\hline
LiPS(Ours) & \textcolor{blue}{\textbf{Low}} & \textcolor{blue}{\textbf{High}} & \textcolor{blue}{\textbf{High}} \\
\hline
Series-to-Parallel & Low & High & Medium \\
\hline
Passive Ankle & High & Low & Low \\
\hline
\end{tabular}
\caption{Comparison of Different Methods}
\label{table:comparison}
\end{table}

(2) Using passive ankles. As Gu et al. attempted in ~\cite{gu2024advancing} with kp=0, kd=10, setting kp to 0 to achieve only passive damping in the ankle joint. This method greatly simplifies the dual-degree-of-freedom performance of the ankle. In actual use, the coupling of the damping and stiffness terms in the parallel ankle leads to a decline in the robot's performance. In our experiments, we found that setting kp to 0 caused the robot to tilt severely, even leading to damage.

(3) Using \textcolor{blue}{LiPS}. The LiPS method uses a serial model during training and a parallel model during deployment. This method fully utilizes the computational speed of current GPU simulation engines during training, significantly increasing the exploration in reinforcement learning. It also ensures that there is no high-frequency analytical solution computation during deployment on the actual robot, greatly reducing the load on the end side. Using our LiPS method, our humanoid robot Tien Kung can perform various robust gaits in practice.

\begin{figure}[h]
    \centering
    \includegraphics[width=\linewidth]{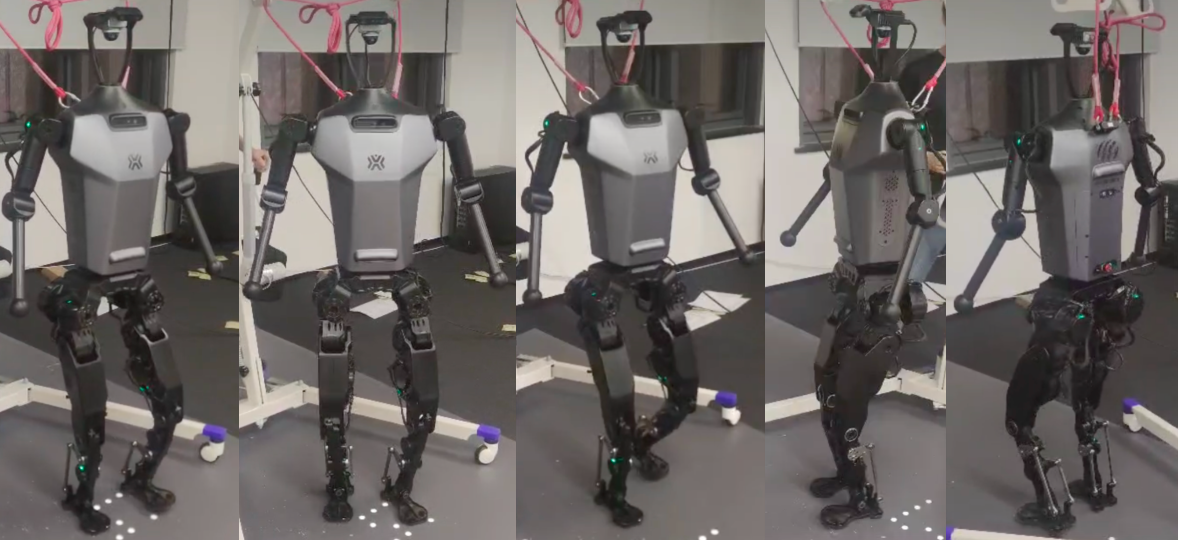}
    \caption{Robust Active Parallel Ankle Omnidirectional Walking Demonstration.}
    \label{fig:para}
\end{figure}

In Fig.~\ref{fig:para}, we demonstrate the robust omnidirectional walking performance of the LiPS method on an actual robot. Due to the significant risk of damage and the varying tuning requirements of other methods, we are unable to present ablation results for all conditions within the limited scope of this paper. However, based on our dynamic modeling and experimental waveform results~\ref{fig:wave}, we can qualitatively conclude that the LiPS method outperforms other methods in terms of computational load, control performance, and joint noise robustness, as shown in Table.~\ref{table:comparison}. Our work introduces a novel perspective and method for the application of series-parallel configurations in humanoid robots within reinforcement learning. We will release our code and framework to community and that can assist researchers to explore new configurations and algorithms, thereby helping the entire community better understand the role of dynamic models in humanoid robotic reinforcement learning.



\bibliographystyle{IEEEtran}
\typeout{}
\bibliography{IEEEabrv,mybibfiles}
\theendnotes
\end{document}